\documentclass[letterpaper, 10 pt, conference]{ieeeconf}  %

\IEEEoverridecommandlockouts                              %

\usepackage{graphics} %
\usepackage{graphicx}
\usepackage{epsfig} %
\usepackage{times} %
\usepackage{amsmath} %
\usepackage{amssymb}  %
\usepackage{multirow}
\usepackage{threeparttable}
\usepackage[font=footnotesize]{subcaption}
\usepackage[font=footnotesize]{caption}
\usepackage{hhline}
\usepackage[official]{eurosym}
\usepackage{balance}
\usepackage{url}
\usepackage{bookmark}
\usepackage[colorinlistoftodos]{todonotes}

\title{\LARGE \bf
OpenRoACH: A Durable Open-Source Hexapedal Platform with Onboard Robot Operating System (ROS)
}

\author{Liyu Wang, Yuxiang Yang, Gustavo Correa, Konstantinos Karydis, Ronald S. Fearing%
\thanks{*This work was supported by the Swiss National Science Foundation Postdoc Mobility Fellowships.}%
\thanks{Liyu Wang, Yuxiang Yang, and Ronald S. Fearing are with Dept. of Electrical Engineering and Computer Sciences,
        University of California Berkeley, CA, USA
        {\tt\small liyu.wang@wadh.oxon.org, \{yxyang, ronf\}@berkeley.edu}}%
        \thanks{Gustavo Correa and Konstantinos Karydis are with Dept. of Electrical and Computer Engineering,
        University of California Riverside, CA, USA
        {\tt\small \{gcorr003, karydis\}@ucr.edu}}%
}

\begin{document}

\maketitle
\thispagestyle{empty}
\pagestyle{empty}

\begin{abstract}
OpenRoACH is a 15-cm 200-gram self-contained hexapedal robot with an onboard single-board computer. To our knowledge, it is the smallest legged robot with the capability of running the Robot Operating System (ROS) onboard. The robot is fully open sourced, uses accessible materials and off-the-shelf electronic components, can be fabricated with benchtop fast-prototyping machines such as a laser cutter and a 3D printer, and can be assembled by one person within two hours. Its sensory capacity has been tested with gyroscopes, accelerometers, Beacon sensors, color vision sensors, linescan sensors and cameras. It is low-cost within \$150 including structure materials, motors, electronics, and a battery. The capabilities of OpenRoACH are demonstrated with multi-surface walking and running, 24-hour continuous walking burn-ins, carrying 200-gram dynamic payloads and 800-gram static payloads, and ROS control of steering based on camera feedback. Information and files related to mechanical design, fabrication, assembly, electronics, and control algorithms are all publicly available on https://wiki.eecs.berkeley.edu/biomimetics/Main/OpenRoACH.
\end{abstract}

\section{Introduction}
Legged robots provide adaptability in environments where the complexity of ground varies, or when continuous contact paths are unavailable~\cite{ROB-044}. Compared to large legged robots (over half a meter body length and over 1 kg weight) that typically cost over tens of thousands of dollars, smaller legged robots may be better suited in applications that require greater accessibility (e.g., to locate survivors trapped under rubble). Comparatively, smaller legged robots can be fabricated quickly and at relatively low costs, often by utilizing only benchtop rapid-prototyping machines (which are nowadays ubiquitous). %
   \begin{figure}[t]
   \vspace{6pt}
      \centering
     \includegraphics[width=0.48\textwidth]{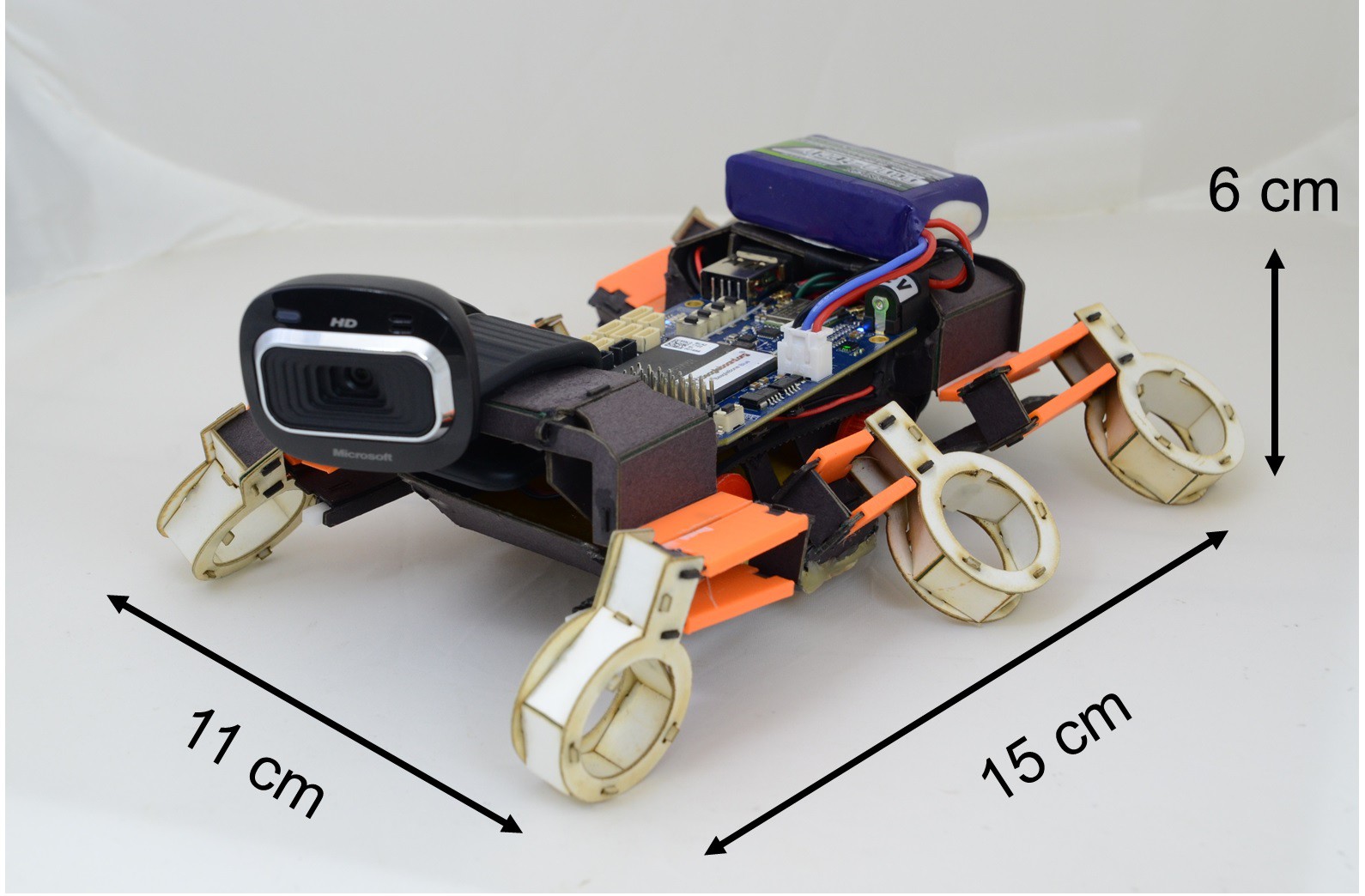}
      \caption{An OpenRoACH with a BeagleBone Blue board, a camera, and a LiPo battery.}
      \label{openroach} 
   \end{figure}

The dimension ranges from several centimeters to just above ten centimeters represents the current technical boundary as how small legged robots can go with actuators, sensors, microcontrollers and energy sources all onboard (robots at the scale of 0.1-10 mm typically rely on external actuators such as magnetic fields or external power sources). These small robots are usually low-cost but with a low payload capacity, which is needed for carrying computing units, additional sensors, and other objects. Table~\ref{smallpayload} shows a comparison of representative legged robots between 5 cm and 20 cm in terms of dimension, mass, and payload capacity. The value of 5 cm was chosen based on the fact that the smallest ROS-capable single-board computer (SBC) at present, NanoPi Neo has a dimension of 4 cm$\times$4 cm.

In addition, open-source legged robots could enable wider research in legged robotics and locomotion and contribute to the overall popularity of legged robotics. We use the definition from the Open Source Robotics Foundation for open-source robotics: development, distribution, and adoption of open-source software and hardware for robotics research, education, and product development~\cite{ROS}. {For open-source robotics hardware we use the definition from the Open Source Hardware Association: be publicly available in a preferred format, use readily-available components and materials, standard processes, open infrastructure, unrestricted content, and open-source design tools~\cite{OSHWA,ORH}.
There are only a handful of fully open-sourced legged robots, such as Marty~\cite{Marty}, ROFI~\cite{ROFI}, Hexy~\cite{HEXI}, Metabot~\cite{7781990}, Aracna \cite{Lohmann2012} and Poppy~\cite{Poppy}. They are all above 30 cm, mostly fully actuated with tens of servomotors, and controlled with an onboard microcontroller (with the bipedal Marty being the only one supporting onboard ROS).

Here we introduce OpenRoACH as the smallest legged platform with onboard ROS (see Fig. \ref{openroach}). OpenRoACH is fully open-sourced. Its hardware computer aided design (CAD) files are publicly available in a preferred format. Its mechanical hardware uses two off-the-shelf motors and accessible materials such as paperboard, nylon, polyethylene terephthalate (PET), and thermal adhesives. It may be fabricated with standard processes such as lamination, laser cutting and fused filament fabrication (FFF) on benchtop fast-prototyping machines. After fabrication, it may be folded and assembled manually within two hours. Its electronic hardware uses off-the-shelf components including a SBC. OpenRoACH software may be developed with a cross-compiler or in ROS installed on the SBC. The overall cost of material for an OpenRoACH configured similarly to the one in Fig. \ref{openroach} was under \$150 (2018), with 80\% of the cost going for electronics and motors. Fig.~\ref{patel} shows the comparative advantage of OpenRoACH over representative legged robots.

\begin{table}[h!]
\vspace{0pt}
\begin{center}
\begin{threeparttable}
	\caption{Comparison of 5-20 cm Legged Robots}
	\label{smallpayload}
	\begin{tabular}{p{2.3cm}|p{1.5cm}|p{1cm}|p{1.8cm}}
		\hline
        Legged robot &Dimension (cm$\times$cm$\times$cm)&Mass$^1$ (g) & Payload$^2$ (g) \\\hline\hline
       Mini-Whegs \cite{Morrey2003} & 9$\times$6.8$\times$7.2 & 146 & 290 (s) \\\hline
       iSprawl \cite{Kim2006} & 15.5$\times$11.65$\times$7 & 300 & N/A \\\hline
       PSR \cite{psari_icra_15} & 18$\times$11$\times$9 & 334 & N/A \\\hline
       SPIDAR \cite{Karydis2017a} & 14$\times$15$\times$6 & 350 & 1000 (s) \\\hline
       Kamigami ~\cite{Kamigami}&11.5$\times$11$\times$4 & 57& 50 (s)\\\hline
       VelociRoACH ~\cite{Haldane2013}&10$\times$6.5$\times$4&37&125 (s)\\\hline
       OpenRoACH&15$\times$11$\times$6 & 178-193 & 800 (s), 200 (d) \\\hline
	\end{tabular}
\begin{tablenotes}
      \small
      \item 1. including electronics and batteries. 2. s: static; d: dynamic.
\end{tablenotes}
\end{threeparttable}
\end{center}
\vspace{-6pt}
\end{table} 

\begin{figure}[h!]
\vspace{-6pt}
      \centering
      \includegraphics[width=0.48\textwidth]{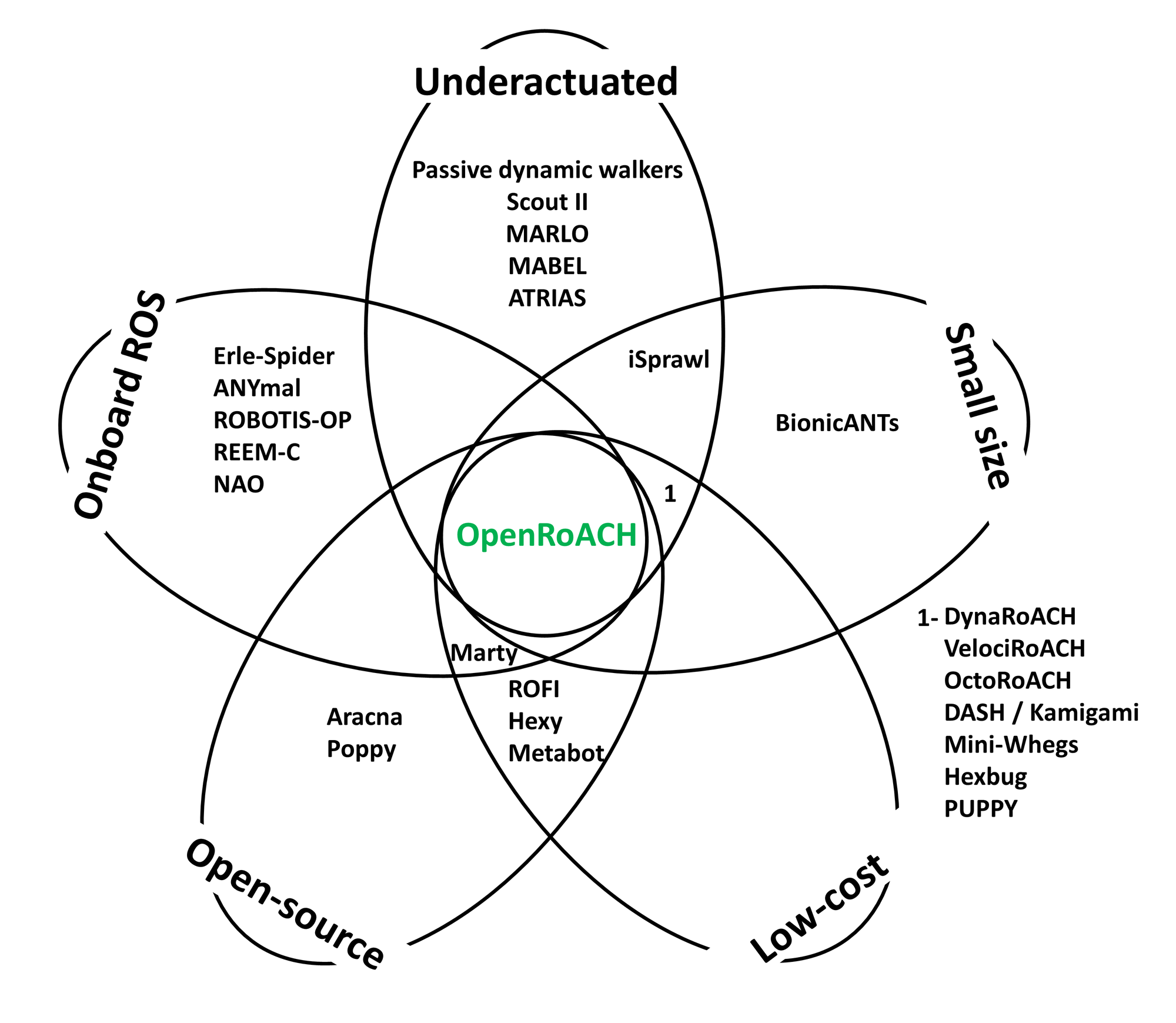}
      \caption{A petal diagram showing the comparative advantage of OpenRoACH among representative legged robots. Low cost: below \$500. Small size: 5-20 cm. Open-source: mechanical hardware, electronic hardware, and software.
      \label{patel}}
\vspace{-6pt}
   \end{figure}

\section{Mechanical Design and Fabrication}
OpenRoACH is folding-based~\cite{Hoover2008} so that minimal assembly is required. Its design relies on careful patterning of creases which, once folded, will become either structure linkages or flexure joints. The mechanical structure consists of a chassis, two leg transmissions, and six legs. Fig.~\ref{cad}(a) shows the CAD sketches of the chassis (top) and the leg transmission (bottom). The chassis consists of four parts, which can be folded and connected to each other by pairs of pegs and holes in designated locations. The leg transmission couples the three legs on each lateral side to the motor module through a transmission bar. Fig.~\ref{cad}(b) shows a partially-assembled OpenRoACH including the chassis and the leg transmission.
\begin{figure}[h]
\vspace{6pt}
      \centering
      \begin{subfigure}{0.22\textwidth}
        \includegraphics[trim=240 100 385 390, clip, width=\textwidth]{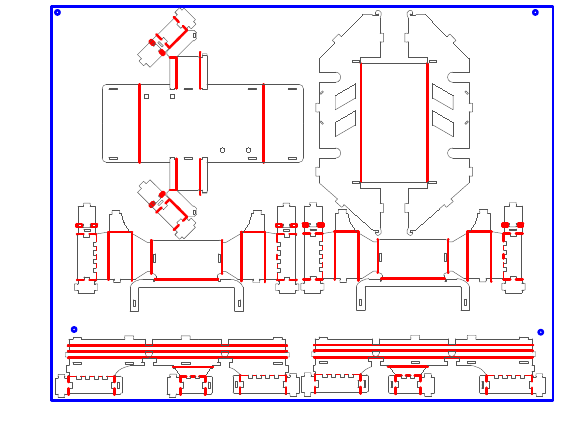}
        \caption{}
      \end{subfigure}
      \begin{subfigure}{0.22\textwidth}
        \includegraphics[width=0.93\textwidth]{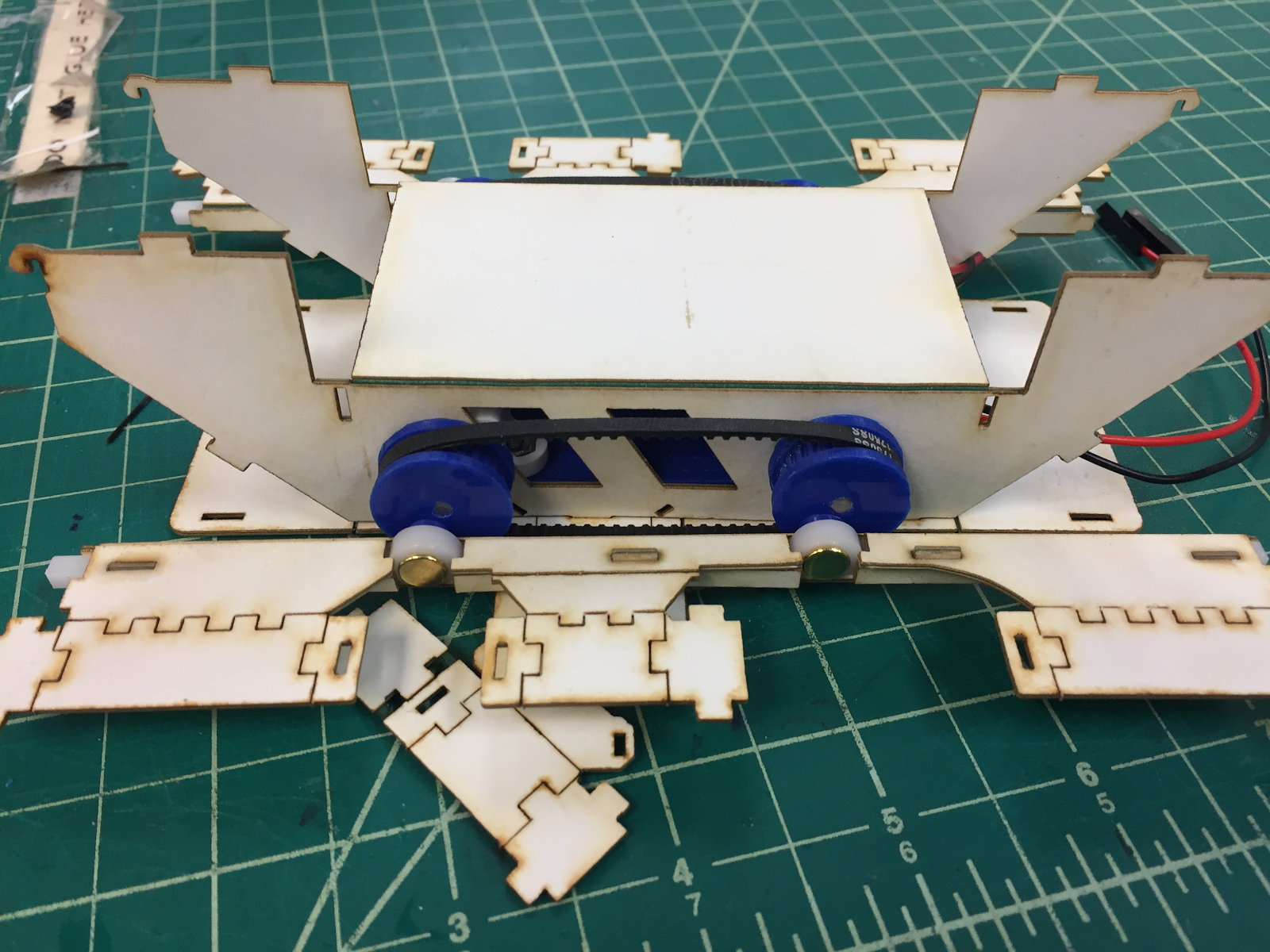}
        \caption{}
      \end{subfigure}
      \caption{(a) CAD drawings of OpenRoACH body, including a chassis (top) and two leg transmissions (bottom) (b) A partially-assembled OpenRoACH.
      \label{cad}}
\end{figure}

\subsection{Principles for Scaling Up SCM}
OpenRoACH's mechanical system weighs 115 gram, and measures 15 cm$\times$11 cm$\times$6 cm. Most of the body parts are fabricated with the smart composite microstructures (SCM) process~\cite{Hoover2008} and then manually folded. The chassis and leg transmissions are scaled up from those of VelociRoACH (see Table~\ref{oldnew}). However, adjustment of dimensions alone~\cite{git} does not scale up the functionality. For example, an attempt with only size scaling failed to even support the larger body mass~\cite{Fitzner2017}. In many other attempts, the flexures quickly suffered from a permanent damage including over-stretch, fracture, tearing, and peeling, or the body had no ground clearance during locomotion. To scale up SCM, three principles have been adopted.
\begin{table}[h]
\begin{center}
\begin{threeparttable}
	\caption{Scaling from VelociRoACH to OpenRoACH}
	\label{oldnew}
	\begin{tabular}{p{3.5cm}|p{1.9cm}|p{1.9cm}}
		\hline
        &VelociRoACH&OpenRoACH \\\hline\hline
        Dimension (cm) &10$\times$6.5$\times$4&15$\times$11$\times$6\\\hline
        Mass structure \& motors (g)& 23 & 115\\\hline
		Mass electronics no battery (g) & 6.1 & 40 \\\hline
        Body structure material & poster-board, thermal adhesive, PET& poster-board, thermal adhesive, PET, nylon, PLA\\\hline
        Fabrication & laser cutting, lamination, casting & laser cutting, lamination, 3D printing \\\hline 
        Structure peel strength (N/m)& 583 & 2167\\\hline
        Maximal motor torque (mNm) & 8.4 & 84-400 \\\hline
        Dynamic payload (gram) & - & 200 \\\hline
		Open sourced&No&Yes\\\hline
		ROS&No&Yes\\\hline
	\end{tabular}

\end{threeparttable}
\end{center}
\end{table}

\subsubsection{Scale Flexure Stiffness, Strength and Yield}
The body parts of OpenRoACH are designed and fabricated with four materials. Posterboard (0.3-0.5 mm thick) is used for the outer layers of non-flexure linkages, ripstop nylon (0.15 mm thick) and PET (0.050-0.075 mm thick) are used for flexure layers, and they are all bonded with thermal adhesives. Fig.~\ref{scm} shows the materials and their arrangement of layers for the relevant body parts. 
\begin{figure}[h!]
\vspace{6pt}
      \centering
      \includegraphics[width=0.48\textwidth]{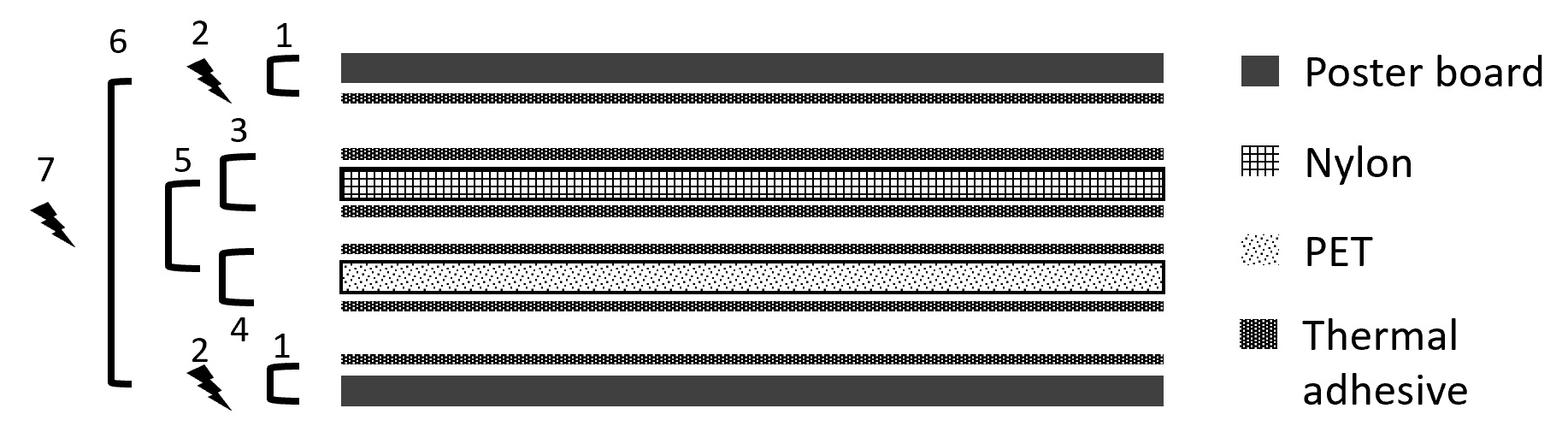}
      \caption{Materials and their arrangement of layers for the body of the robot. The SCM fabrication process use laminating (steps 1, 3-6) and laser cutting (steps 2, 7) alternatively.
      \label{scm}}
      \vspace{-6pt}
   \end{figure}

Different from previous applications of the SCM process to manufacture similar legged robots, OpenRoACH uses two materials (that is, ripstop nylon and PET) for the flexure layer instead of just one. Since ripstop nylon and PET have comparable modulus and strength, their combination can help maintain both the flexure stiffness and its strength. Practically, the dual-material approach prevents flexures from getting permanently over-stretched as compared to using a thin layer of ripstop nylon alone, and prevents fractures as compared to using PET alone.

\subsubsection{Improve Balance and Weight Support}
To balance a longer body and support larger body mass, a motor module was designed to have a front and a rear shaft on each lateral side of the robot. The overall motor module consists of two DC motors, two passive shafts, two motor mounting brackets, a motor mounting base, four pulley-flanges, and two timing belts. On each lateral side of the robot, a pulley-flange is mounted on the DC motor shaft and the passive shaft respectively. The two shafts, the leg transmission, and the two pulley-flanges form a four-bar linkage and ensure the leg transmission is parallel to the bottom of the robot. Timing belts are added to avoid anti-parallelogram locks and ensure the synchronization between the motion of the two pulley-flanges. The DC motors (micro metal gearmotors, Pololu, USA), motor mounting brackets (Pololu, USA), and timing belts (SDP-SI, USA) are all off-the-shelf components. The motor mounting base and the pulley-flanges are designed for 3D printing. %
   
Fig.~\ref{mechsys} highlights several key mechanical components. 
Fig.~\ref{mechsys}(a)-(b) show the top internal and front views of assembled motor module situated in the body as well as the side view of the dual shafts with a timing belt. 
Fig.~\ref{mechsys}(c) shows CAD sketches of the pulley-flange for the motor shaft (left) and that for the passive shaft (middle).
Fig.~\ref{mechsys}(d) shows a CAD sketch of the motor mounting base. The motor is mounted by the brackets while the passive shaft rotates freely within the tunnel. 
\begin{figure}[h!]
   \vspace{6pt}
      \centering
  \includegraphics[width=0.465\textwidth]{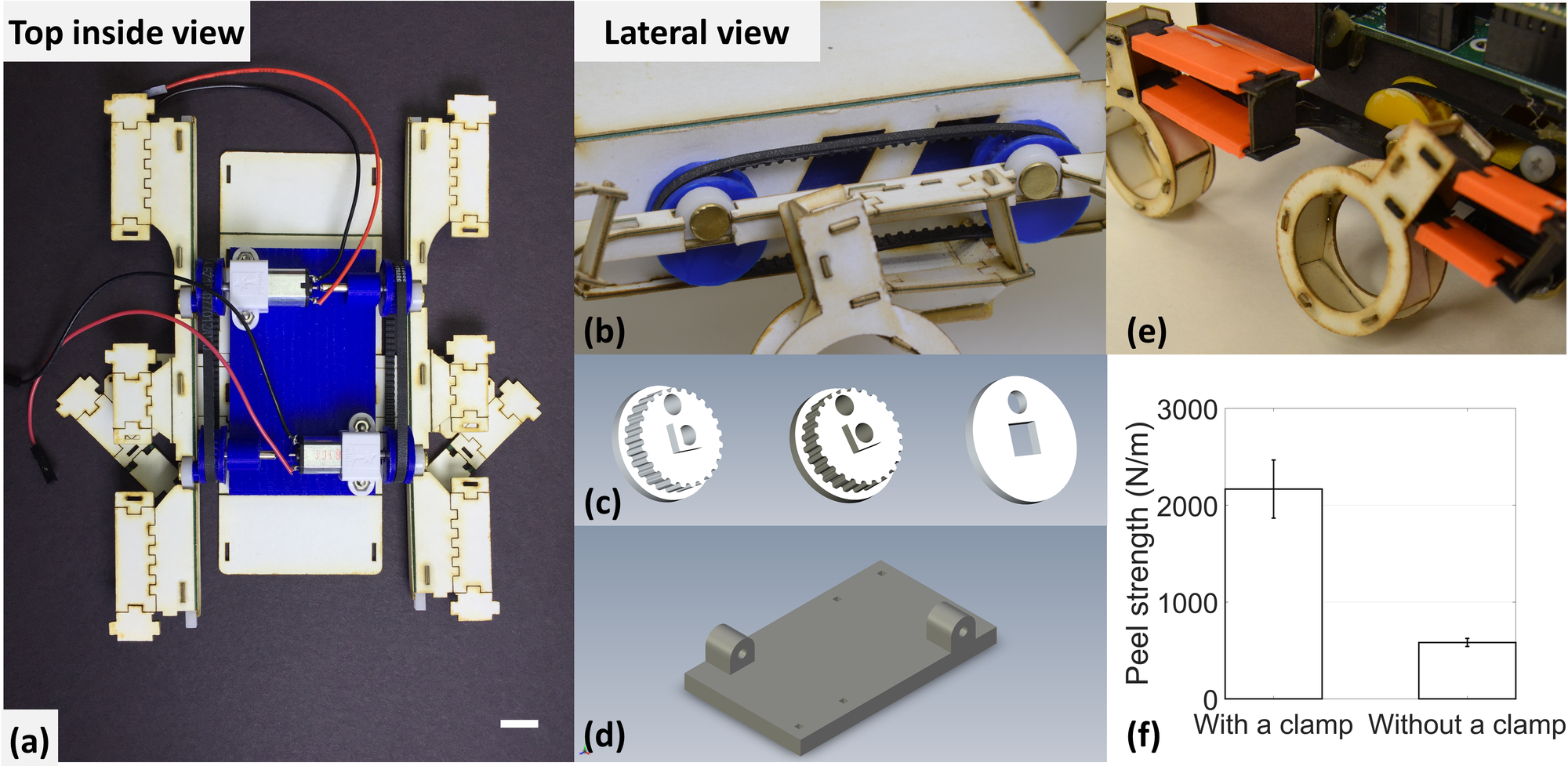}
      \caption{(a) Top inside view of a partially assembled OpenRoACH mechanical system. The white bar indicates 1 cm. (b) Lateral view of the dual-shafts, pulley-flanges, the leg transmission bar, and the timing belt. (c) and (d) CAD sketches of OpenRoACH's pulley-flange and motor base. (e) Reinforcing PLA clamps on the leg transmission. (f) Peel strength of structures with and without the clamps (from 5 trials each).
      \label{mechsys}}
      \vspace{-12pt}
   \end{figure}

\subsubsection{Increase Peel Strength}

To increase the peel strength between the flexure layers and outer layers to accommodate the mass increase, clamps may be added to the non-flexure part of the structure. Clamps were chosen over other reinforcement methods such as rivets because they offer a non-invasive solution to the existing mechanical design. 
Based on the existing mechanical design as shown in Fig.~\ref{cad}, three dimensions of clamps were designed and 3D printed (Fig.~\ref{mechsys}(e)). A total of 18 clamps are needed for an OpenRoACH, all in leg transmissions.

Peel strength was tested to quantify the effect of these clamps. Tests were carried out on a mechanical strength testing machine (Instron, USA) with an extension rate of 1 mm/s. Three samples with a width of 12 mm were tested for cases with and without a clamp made of PLA. 
Results show mean peel strengths of 2167 N/m and 583 N/m for the cases with and without a clamp, respectively (Fig.~\ref{mechsys}(f)).

\subsection{Fabrication and Assembly}
The entire mechanical system of the OpenRoACH can be fabricated with benchtop fast-prototyping machines. The body is fabricated with a laminator and a laser cutting machine (e.g., Versa LASER 200). The SCM process alternates between the laminator and the laser cutter to make sandwich structures with flexures from layers of materials (see Fig.~\ref{scm}). In addition, the leg transmission bar is laser cut from a hard but not brittle material such as polyoxymethylene (POM or Delrin). The motor module and the reinforcing clamps are fabricated with a 3D printer. For the motor module, when the 0.25 mm diameter nozzle is used on a Ultimaker 2+, it would take 11-12 hours to fabricate the motor mounting base with 75\% fill density and 1.5 hours to fabricate a single pulley-flange with 50\% fill density.

Manual assembly is required to fold and connect the laser-cut body parts as well as to put the motor module together. 
Assembly time is approximately two hours for an engineering bachelor student or a senior high school student with engineering knowledge.

\section{Electronics and Control}
OpenRoACH's electronic systems can be flexibly configured between using a microcontroller and/or a single board computer (SBC), all of which can be acquired off the shelf. 
Three configurations of the electronic system have been tested on OpenRoACH.  These include i) a BeagleBone Blue board, ii) a Raspberry Pi 3 combined with an mbed microcontroller, and iii) a standalone mbed microcontroller. BeagleBone Blue offers the lightest-weight (40 grams) and the most convenient solution, while the latter two require additional components such as a motor driver. We have developed an open-source carrier board that mounts a motor driver carrier board (DRV8833, Pololu, USA) and an mbed microcontroller (NXP LPC1768, ARM, UK). The configuration of an mbed with the carrier board weighs 58 grams; that of a Raspberry Pi 3, an mbed, and the carrier board weighs 105 grams. In all configurations, OpenRoACH is powered by a 7.4V LiPo battery.

Depending on the electronic system configuration, there are two ways to control OpenRoACH. When an mbed is used, control algorithms are developed in C++ in the mbed online compiler running on a PC and then downloaded to the microcontroller as binaries. When ROS is used, information about OpenRoACH is published as a ROS topic, such that other ROS packages could subscribe to OpenRoACH to read sensor data and publish control commands.

A variety of sensors have been tested for feedback control of OpenRoACH locomotion. These include gyroscopes, accelerometers, Beacon sensors, color vision sensors, linescan sensors, and cameras. For example, accelerometers (e.g. MPU-6050) have been used for control of motor on/off based on vertical acceleration caused by a natural fall or pick-up by a robot arm; Beacon sensors (e.g. Pololu IR) have been used for a homing task within the range of 0.2-2 meters. For the use of linescan sensors (e.g. TSL1401, Parallax, USA) and cameras, please refer to Section IV.D and IV.E respectively.

\section{Experiments and Evaluation}
The performance of OpenRoACH has been evaluated in five experiments: multi-surface locomotion, durability, payload, and motion tracking with line and web cameras.

\subsection{Multi-Surface Walking \& Running (mbed)}
Walking and running is tested in open loop on four types of surfaces: flat smooth tile surface, flat smooth wood surface, a flat rough carpet surface, and a rough terrain of pebble gravel. In these tests, OpenRoACH is equipped with an mbed microcontroller and carrier board (total 58g), and is powered by a 23g LiPo battery. Two speed settings were realized by using motors with different gear ratios: for the relatively low speed a 298:1 ratio was used for 100 rpm, while for the relatively high speed a 50:1 ratio was used for 625 rpm. 

Fig.~\ref{walkrun} shows sagittal-plane snapshots of running with relatively high speed on the four surfaces. Open-loop walking was achieved with the relatively low speed on the four surfaces too. Table~\ref{walkruntable} lists average forward speed of the OpenRoACH running and walking from three trials under each condition. With both gait patterns, OpenRoACH moved faster on wood board than on tile and on pebble gravel. What is interesting is the case of operating on carpet; in running the robot had the highest average forward speed but in walking it was only slightly faster than on pebble gravel. This may be due to slipping of legs in contact with tile and wood board in the case of running.
\begin{figure}[t]
\vspace{6pt}
	\centering
	\includegraphics[width=0.48\textwidth]{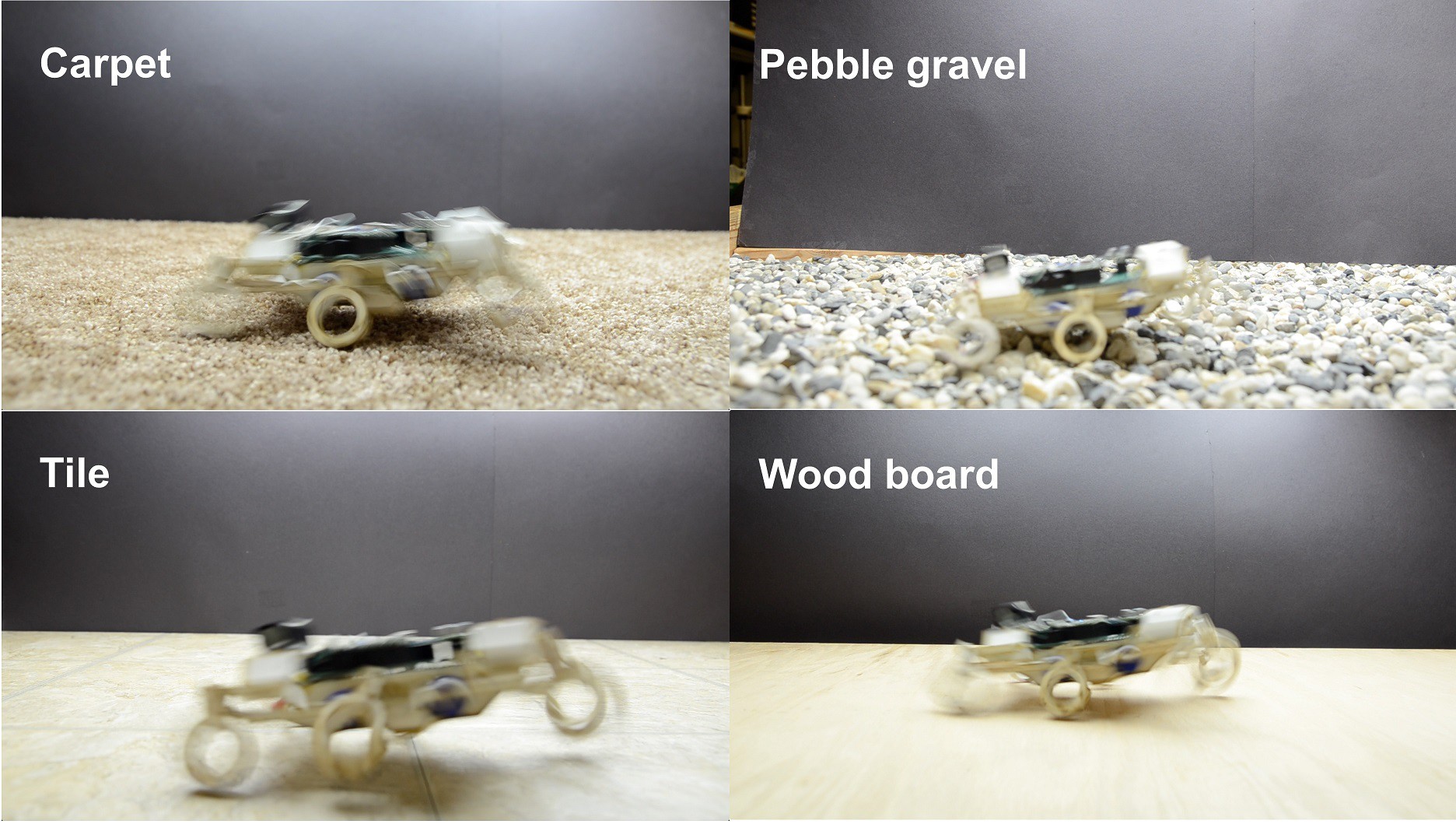}
	\caption{Open-loop running on different surfaces or terrain.}
\label{walkrun}
\end{figure}

\begin{table}[t]
\vspace{6pt}
	\centering
	\caption{Results on Ground Surfaces or Terrain}
	\begin{tabular}{c|c|c|c|c}
		\hline
		\multirow{2}{*}{} &\multicolumn{2}{c|}{Running}&\multicolumn{2}{c}{Walking}\\
		\hhline{~----}
		 & Speed (m/s) & Freq (Hz) & Speed (m/s) & Freq(Hz)\\\hline\hline
		Carpet&0.343&8&0.0854&2\\\hline
		Pebble gravel&0.117&9.5&0.0650&2.1\\\hline
		Tile&0.245&10&0.1104&2.1\\\hline
		Wood board&0.284&8&0.1231&1.8\\\hline
	\end{tabular}
	\label{walkruntable}
    \vspace{-6pt}
\end{table}

\subsection{Walking Burn-In Tests (mbed)}
Two walking burn-in tests were carried out for two OpenRoACHes to test mechanical durability and overall reliability. Fig.~\ref{burnin} shows the burn-in setup, which contains a treadmill whose surface is made from canvas fabric. To keep the robots from falling off from the front or the rear edges, a finite-state feedback controller was implemented where the robots adjusted their motor speed based on the linescan sensor detection of a static black-white edge. Both OpenRoACHes were tethered to an external power supply through a current sensor (ACS712) for power measurement. Both robots are equipped with the mbed microcontroller and carrier board.
\begin{figure}[h!]
	\centering
	\includegraphics[width=0.45\textwidth]{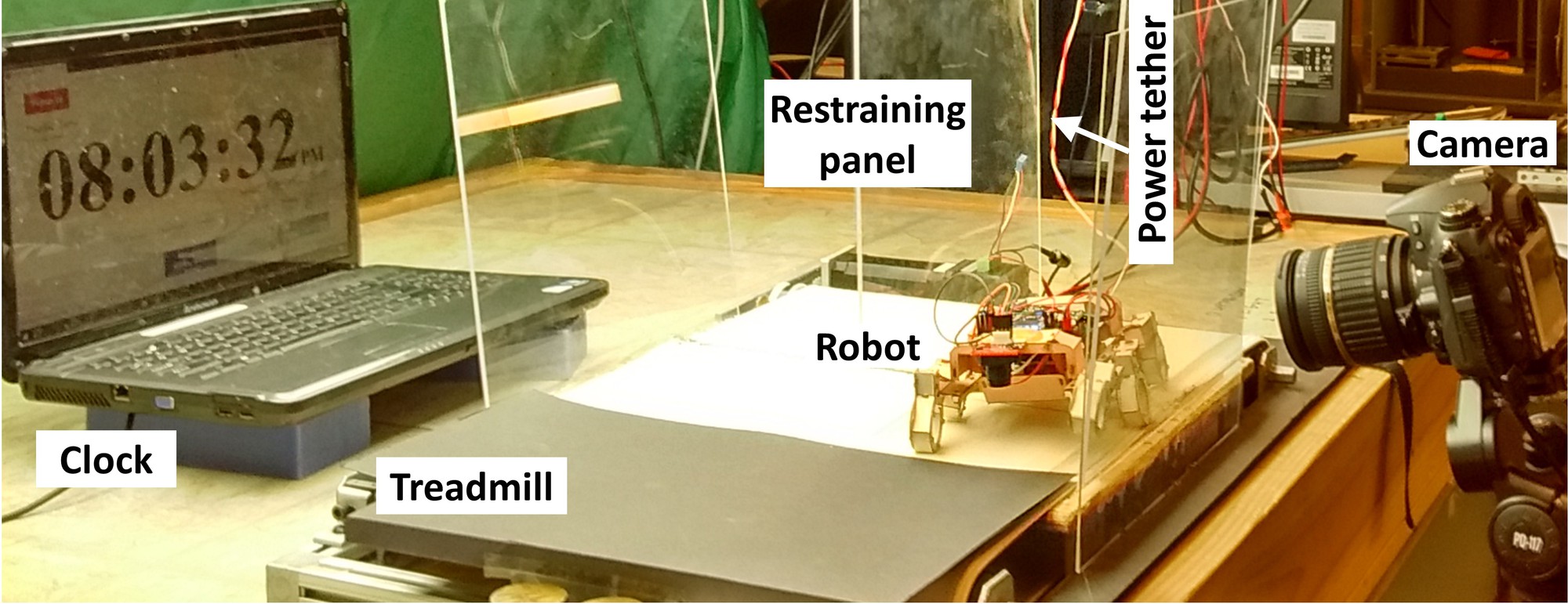}
	\caption{Burn-in test setup.}
\label{burnin}
\vspace{-6pt}
\end{figure}

In the first burn-in test, an OpenRoach without the clamps walked over 24 hours continuously with a voltage supply of 7.67 V. The speed of the treadmill was 0.06 m/s for the first six hours and 0.03 m/s for the rest of the test. Walking was mostly continuous except for three major interruptions due to motor and pulley flange replacement at 11 hours 19 minutes, 16 hours 00 minutes, and 22 hours 48 minutes. Mechanical deterioration happened mainly on the leg transmission with the middle hip suffering modest peeling and the front and rear hips suffering minor peeling. 

Table \ref{burnintable} shows kinematic data before and after the burn-in; the robot was tasked to walk in open loop for ten seconds on the flat wood surface. Three trials were recorded for each case via motion capture. The mean and the standard deviation (STD) values were calculated over all the strides in a single trial. The mean value of the average height of the robot decreased from 0.062-0.071 m to 0.052 m, or by 16.1\%-26.8\%. In terms of pitch angle, the mean value changed from positive to negative, indicating the front of the robot tended to point downwards after burn-in. Both the STD of average height and pitch angle increased after the burn-in test.
\begin{table}[h]
	\centering
	\caption{Results Before and After a 24-hour Burn-In}
	\begin{tabular}{c|c|c|c|c|c}
		\hline
		\multirow{1}{*}{} &\multirow{1}{*}{Trial ID}&\multicolumn{2}{c|}{Before}&\multicolumn{2}{c}{After}\\
		\hhline{~~----}
		& & Mean & STD & Mean & STD\\\hline\hline
         
\multirow{ 3}{*}{Average height (m)} & 1& 0.071 & 0.002 & 0.052 & 0.004 \\
&2& 0.070 & 0.002 & 0.052 & 0.003  \\
&3& 0.062 & 0.002 & 0.052 & 0.003  \\ \hline       
\multirow{ 3}{*}{Pitch angle (degree)} & 
1&1.475 & 1.705 & -3.969 & 4.390 \\
&2& 2.091 & 2.386 & -3.304 & 5.070  \\
&3& -0.915 & 2.029 & -4.380 & 4.935  \\ \hline 
	\end{tabular}
	\label{burnintable}
\end{table}

In the second burn-in test, another OpenRoACH walked accumulatively for more than five hours with a power supply of 10.68 V. The speed of the treadmill was 0.0384 m/s. The robot had reinforcing clamps on its hips and leg transmission. Fig.~\ref{burnincurrent} shows the power consumption by the two motors over the five hours, measured by a current sensor (ACS712) at 5 Hz. It can be seen the power decreases over time. The average power consumption for the five hours were 3.77 W, 3.90 W, 3.38 W, 3.01 W and 3.06 W, respectively, indicating a 18.8\% decrease in average power consumption. This may be due to the reduction of internal friction in the flexures.
\begin{figure}[h]
\vspace{-6pt}
	\centering
	\includegraphics[width=0.51\textwidth]{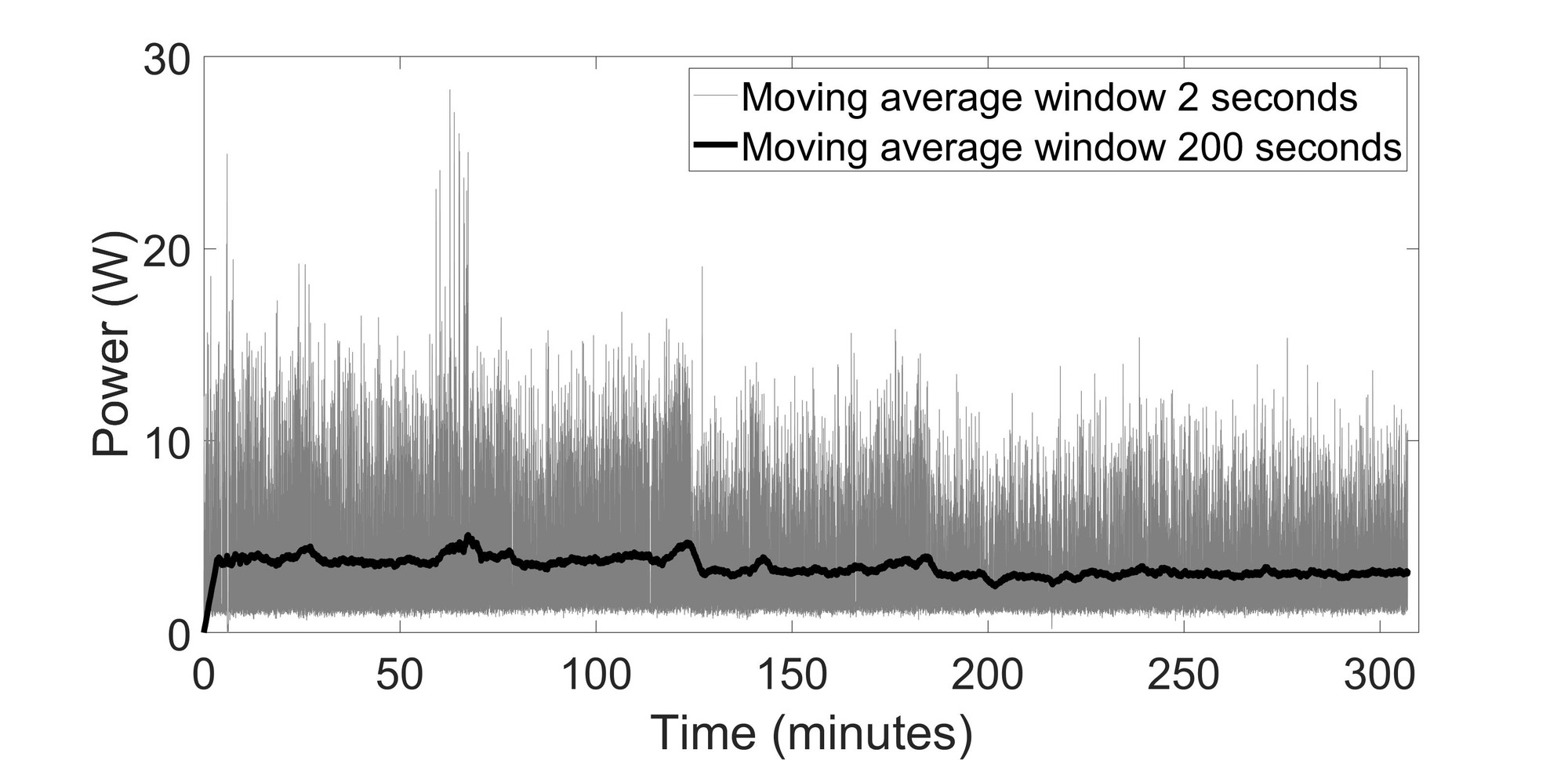}
	\caption{Power data from the five-hour accumulative burn-in walking test. Voltage was at 10.68 V from a power supply.
\label{burnincurrent}}
\vspace{-6pt}
\end{figure}

\subsection{Payload Capacity (mbed \& off-board ROS)}
OpenRoACH enjoys significant payload capacity (better than 1:1) relative to its scale, as indicated via static and dynamic testing. Here we consider the battery as part of the payload. In the tests, the robot (configured with an mbed and carrier board) is powered from a 850mAh 7.4V LiPo battery (58 gram), and is retrofitted with a custom-made 3D-printed payload carrier (29 gram). Three retroreflective markers (1 gram combined) are added for the purpose of collecting data via motion capture. 

In dynamic payload capacity testing, mass is placed on the robot until the first failure. We consider four cases studies with added mass as follows: i) 88 gram (i.e. the combined mass of the payload carrier, battery, and markers), ii) 150 gram, iii) 200 gram, and iv) 250 gram.
For each case study we collected 20 open-loop trajectories each lasting for 10 seconds under constant commanded motor PWM signals. All tests are conducted on a carpeted floor. The robot was able to complete all 88, 150, and 200 gram payload trials with ease, and 18 out of 20 trials with 250 gram payload before the left side motor burned out.
\begin{figure}[t]
	\begin{subfigure}{.495\textwidth}
		\hspace{-6pt}
	\includegraphics[trim={1.5cm 0 2.5cm 0},clip,width=1.0\textwidth]{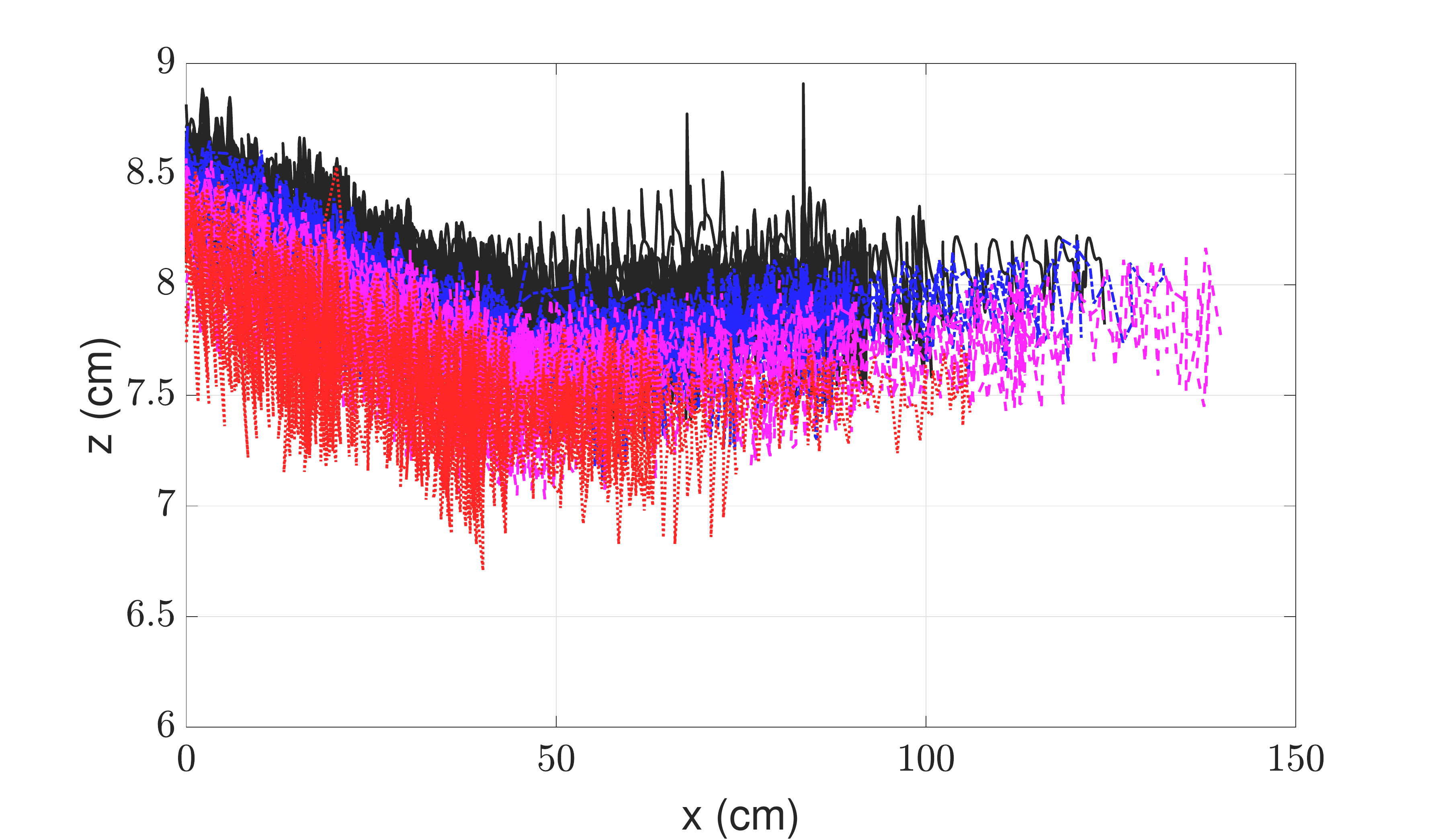}
\end{subfigure}
	\caption{Observed robot trajectories during dynamic payload testing. Solid (black), dashed-dotted (blue), dashed (magenta) and dotted (red) curves correspond to 88, 150, 200 and 250 gram payload, respectively. (Figure best viewed in color.)}
	\label{fig:dynamicPayloadGraphs}
\end{figure}

Individual experimental trajectories are shown in Fig.~\ref{fig:dynamicPayloadGraphs}. 
The height reported corresponds to the plane defined by the three markers, being parallel to the ground plane at an offset of 8.5 cm. Small fluctuations around this value may be observed since the initialization of feet leads to slightly different robot postures. As more weight is added, the robot's height drops less than 0.5 cm on average. Then, when the robot starts moving, it manages to self-regulate its height without any explicit synchronization between the legs on each side, and no matter the payload it is carrying.

As the payload increases, the open-loop control trajectory endpoints demonstrate shorter distances traveled along the X-axis, and are shifted toward the positive Y-axis (leftward), with smaller dispersion. Further, the maximum height difference at final time remains less than 0.5 cm on average.

It is worth mentioning that other than the burned out motor, no other damage was observed after this set of experiments. For static payload capacity testing, mass is placed on the robot until sagging occurs over 50\% of the nominal height. The robot's static payload capacity is at least 800 grams, depending on how the legs are initialized. This point further shows that the robot chassis is durable, as the burn-in and dynamic payload tests demonstrate.

\subsection{Line Following Tracking Test (mbed)}

Further, we demonstrate that OpenRoACH can successfully follow a line based on the line scan sensor's feedback. The robot was tasked to follow a standard figure-8 path, which often serves as a benchmark for motion control. In this case, the robot is configured with only the mbed microcontroller and carrier board. A proportional control was implemented where the difference in the speed of the left and right motors is directly proportional to the detected deviation of the edge of the line from the center of the line scan sensor array. If the sensor ever fails to detect a line, OpenRoACH follows the last known line position. 

Fig.~\ref{8traj}(a) shows the actual trajectory of the OpenRoACH following the figure-8 path made from a 1 cm wide black tape. The trajectory was obtained with a motion capture system (OptiTrack, USA) with four markers placed on the robot. OpenRoACH followed the figure-8 path for three cycles, clockwise on the right half and counterclockwise on the left half. The Root Mean Square (RMS) value of its deviation from the central line of the path is $4.269$ cm. Such a deviation is expected because the motion capture system tracks OpenRoACH's center of mass, while the linescan sensor is placed in front of the robot. The more volatile deviation in right-turning (see Fig.~\ref{8traj}(b)) is due to the lateral asymmetry in individual motors and fabrication and assembly process. Four other trials with three cycles were achieved successfully.
\begin{figure}[h]
\vspace{-0pt}
	\centering
  	\includegraphics[width=0.485\textwidth]{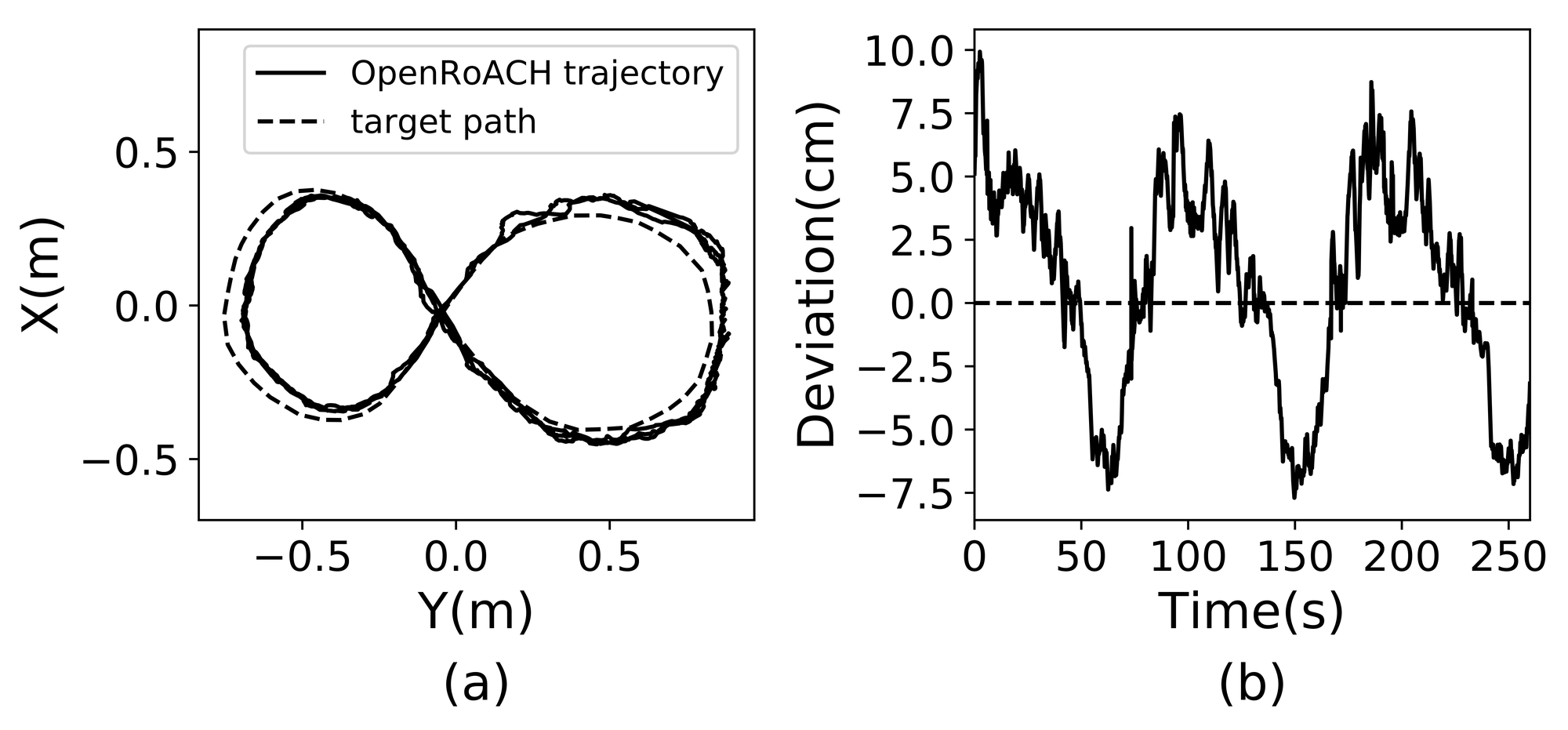}
	\caption{Line-following results of an OpenRoACH. (a) Trajectory (in blue) of OpenRoACH following a figure-8 path (in green) for three cycles. (b) Deviation from the line center.
\label{8traj}}
\end{figure}

\vspace{-16pt}
\subsection{AR Tag Detection and Tracking (BeagleBone Blue \& onboard ROS)}
In the last set of experiments, we show that OpenRoACH is capable of tracking an AR tag using an onboard USB webcam. Here, the robot is tasked to follow an AR tag on a treadmill (see Fig. \ref{ROS}). The OpenRoACH carried a BeagleBone Blue, a LiPo battery, and USB camera. The onboard camera detected and tracked a moving AR tag at 4 Hz located in front of the robot. A second camera (not shown) was used to track AR tag absolute position in the world frame. A controller was implemented where the difference in the speed of the two motors is directly proportional to the detected deviation of the robot to the center of the AR tag location. %
\begin{figure}[h!]
\vspace{6pt}
	\centering
	\includegraphics[width=0.48\textwidth]{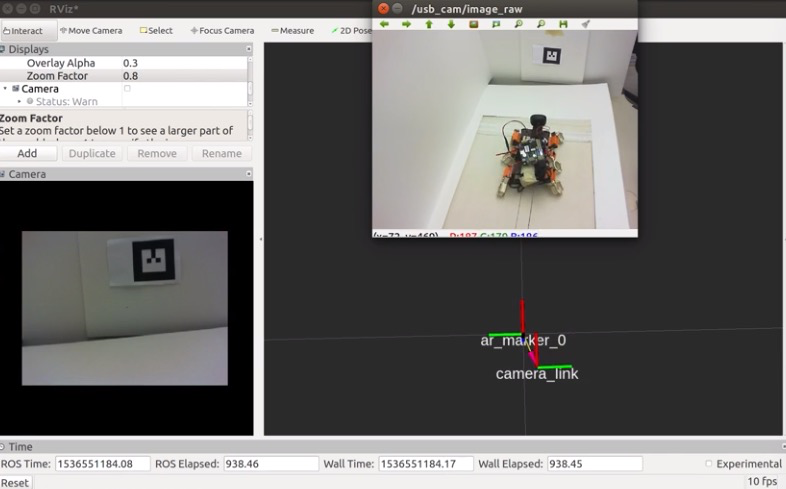}
	\caption{OpenRoACH's camera feedback of the AR tag and a PC interface to visualize operations with onboard ROS.}
	\label{ROS}
    \vspace{-6pt}
\end{figure}

Fig.~\ref{arpos} shows the relative lateral position of the AR tag viewed by the robot's onboard camera (solid line) and the absolute lateral position of the AR tag measured by the second camera (dashed line). It can seen that for the ten changes of the AR Tag's absolute position, the OpenRoACH consistently reacted to reduce relative error, albeit with some delay. Overall, the robot tracked AR tag position with a mean relative error of 0.011 m. 
\begin{figure}[t]
	\centering
	\includegraphics[trim={5cm 0 0 0},clip,width=0.535\textwidth]{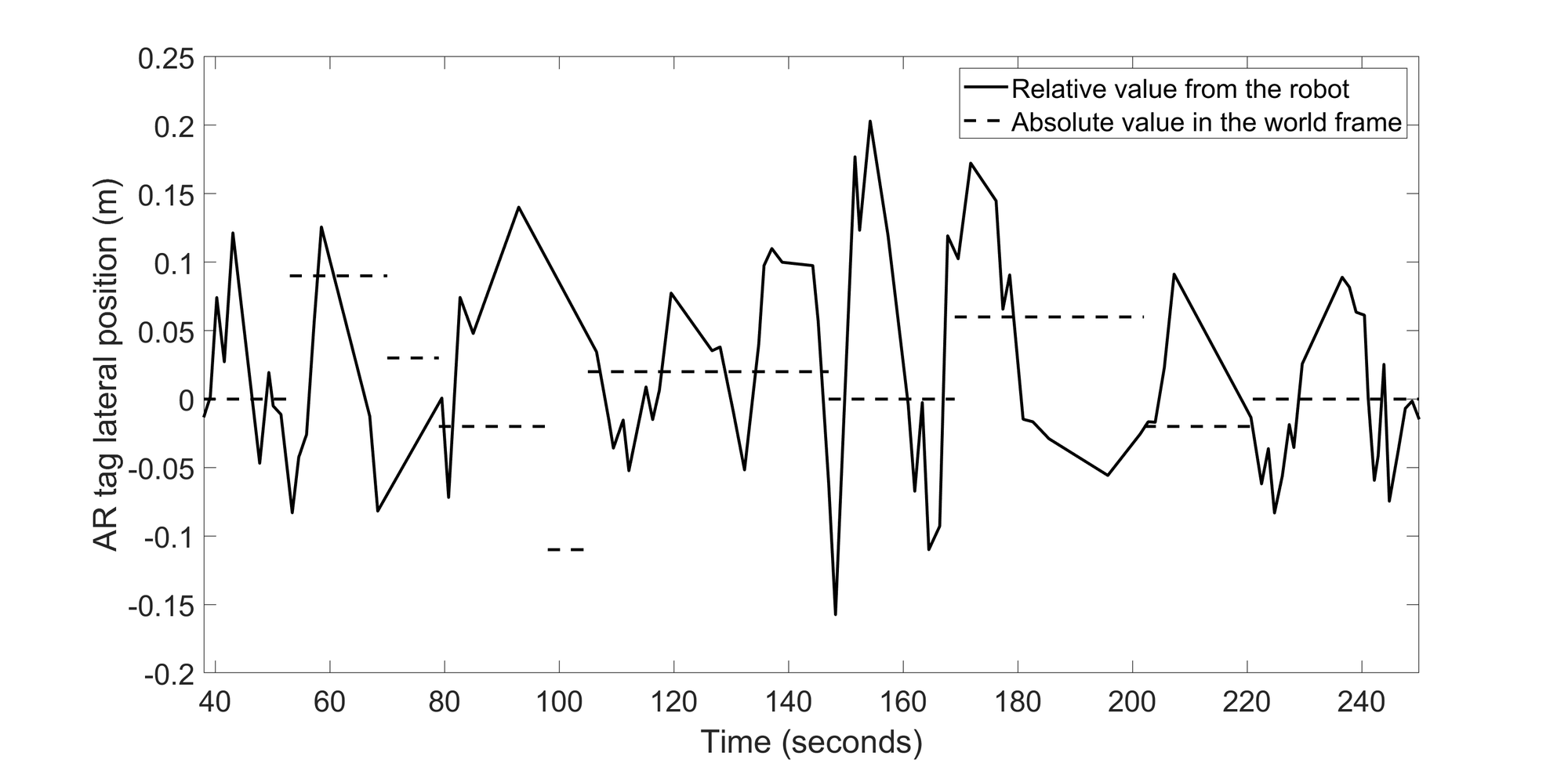}
	\caption{AR tag absolute lateral position in the world frame and relative lateral position from the OpenRoACH's camera.}
	\label{arpos}
    \vspace{-8pt}
\end{figure}

\section{Conclusions}
The paper presents OpenRoACH as a small hexapedal robot with onboard ROS. Scientifically, the platform serves as a validation for three design principles to scales up SCM-based legged robots: (1) scale flexure stiffness, strength and yield; (2) improve balance and weight support; and (3) increase peel strength between flexure and non-flexure layers.

Technically, OpenRoACH offers the following features: (1) fully open-sourced mechanical design, electronics, and software; (2) under-actuated with two DC motors actuating six legs; (3) folding based design; (4) laser cutting and 3D printing enabled fabrication; (5) ROS enabled by a single-board computer; (6) low cost of \$150; and (7) short fabrication and assembly time. It has demonstrated reliable walking on flat ground surfaces with a high gear ratio and multi-surface running with a low gear ratio. It survived a 24-hour continuous walking burn-in. It can carry 200 gram payloads dynamically and 800 gram payloads statically. It has been tested with integrated feedback from a variety of sensors including gyroscopes, accelerometers, Beacon sensors, color vision sensors, linescan sensors and cameras. All of these suggest OpenRoACH is a suitable research tool for legged robot design, fabrication, gait control, and locomotion. For example, it can be used to study sensorimotor learning strategies on various surfaces and terrain, or in the event of damage to one or more of the legs.

Information and files related to mechanical design, fabrication, assembly, electronics, and control algorithms are publicly available on \url{https://wiki.eecs.berkeley.edu/biomimetics/Main/OpenRoACH}.

\addtolength{\textheight}{-0cm}

\section*{Acknowledgments}
We thank A. Fearing, D. Haldane, B. Hogg, S. Liu and other team members for the development of OpenRoACH, and Prof. L. Whitcomb of JHU for assistance with BeagleBone Blue. We also thank T. Lam, S. Gopalakrishnan, and K. Ye for their help in payload testing data collections.

\end{document}